\documentclass[runningheads]{llncs}

\usepackage{eccv}

\usepackage{eccvabbrv}

\usepackage{graphicx}
\usepackage{booktabs}

\usepackage[accsupp]{axessibility}  

\usepackage[pagebackref,breaklinks,colorlinks,citecolor=eccvblue]{hyperref}

\usepackage{orcidlink}

\usepackage{times}
\usepackage{graphicx}
\DeclareGraphicsExtensions{.pdf,.png,.jpg}
\usepackage{amsmath}
\DeclareMathOperator*{\argmax}{argmax} 
\DeclareMathOperator*{\argmin}{argmin} 
\usepackage{amssymb}  
\usepackage{pifont}
\newcommand{\xmark}{\ding{55}}%
\usepackage{makecell}
\usepackage{booktabs}
\usepackage{multirow}
\usepackage{here}
\usepackage{kotex}
\usepackage{soul, color}

\begin{document}

\title{RGBD GS-ICP SLAM} 


\author{Seongbo Ha\orcidlink{0009-0007-7018-1598} \and
Jiung Yeon\orcidlink{0009-0004-7649-549X} \and
Hyeonwoo Yu\orcidlink{0000-0002-9505-7581}}
\authorrunning{Seongbo Ha et al.}

\institute{Sungkyunkwan University, Suwon, South Korea\\
\email{\{sobo3607,wcr12st,hwyu\}@skku.edu}\\
}
\maketitle


\begin{abstract}

Simultaneous Localization and Mapping (SLAM) with dense representation plays a key role in robotics, Virtual Reality (VR), and Augmented Reality (AR) applications. Recent advancements in dense representation SLAM have highlighted the potential of leveraging neural scene representation and 3D Gaussian representation for high-fidelity spatial representation. In this paper, we propose a novel dense representation SLAM approach with a fusion of Generalized Iterative Closest Point (G-ICP) and 3D Gaussian Splatting (3DGS). In contrast to existing methods, we utilize a single Gaussian map for both tracking and mapping, resulting in mutual benefits. Through the exchange of covariances between tracking and mapping processes with scale alignment techniques, we minimize redundant computations and achieve an efficient system. Additionally, we enhance tracking accuracy and mapping quality through our keyframe selection methods. Experimental results demonstrate the effectiveness of our approach, showing an incredibly fast speed up to 107 FPS (for the entire system) and superior quality of the reconstructed map.
\newline\small The code is available at: \href{https://github.com/Lab-of-AI-and-Robotics/GS_ICP_SLAM}{https://github.com/Lab-of-AI-and-Robotics/GS-ICP-SLAM}
\newline\small Video is: \href{https://youtu.be/e-bHh_uMMxE}{https://youtu.be/e\-bHh\_uMMxE}
\end{abstract}

\section{Introduction}
Visual Simultaneous Localization and Mapping (SLAM) is an algorithm that constructs maps of unknown environments while localizing poses of vision sensors simultaneously. As the influence of 3D visual SLAM in the fields of robotics, Virtual Reality (VR), and Augmented Reality (AR) has increased, higher rendering performance and more accurate trajectory are required. Thus, 3D reconstruction methods such as Signed Distance Field (SDF), and Truncated Signed Distance Field (TSDF) are utilized for traditional dense visual SLAM \cite{kinectfusion, dai2017bundlefusion, prisacariu2017infinitam, bylow2013real, whelan2013robust, canelhas2013sdf}. 

Recently, coordinate-based 3D Implicit Neural Representation (INR) \cite{mildenhall2021nerf} has been proposed for representing spatial information using neural radiance fields, showcasing high novel view synthesis capabilities and high-fidelity spatial representation prowess.
Various approaches \cite{imap, voxfusion, nice-slam, pointslam, orbeez, kong2023vmap} have been attempted to utilize INR for the real-time SLAM mapping process.
However, INR requires computationally intensive raycasting to synthesize images, thus the rendering process in INR-based SLAM incurs significant time overhead, slowing down map optimization and rendering-loss based tracking.
In contrast, coordinate-based 3D explicit representation such as 3D Gaussian Splatting (3DGS) \cite{gaussiansplatting} represents the 3D space using 3D Gaussians as primitives, allowing for rendering speeds faster than NeRF using the rasterization method \cite{wu20234d}.
This rapid rendering capability of 3DGS results in fast optimization of 3D spatial information, making it suitable for dense representation in SLAM \cite{gsslam, photoslam, splatam, gaussiansplattingslam}.

Although 3DGS-based SLAM methods take advantages of the high-speed rendering, they fail to address the fundamental issue: the inability to directly utilize 3D explicit representations and the indirect tracking of 3D space through 2D image rendering.
Even with the majority of current 3DGS-based SLAM \cite{pointslam,gsslam,photoslam,splatam,gaussiansplattingslam} utilizing RGB-D data, the use of explicit representations is overlooked.
For example, INR-based methods \cite{imap, nice-slam, pointslam} and 3DGS-based methods \cite{gsslam, splatam, gaussiansplattingslam} employ photometric-error based techniques, which estimate the optimal pose by iteratively minimizing the error between rendered and observed 2D images.
However, due to limitations in tracking speed and performance, decoupled approaches \cite{orbeez, kong2023vmap, photoslam} have been proposed incorporating well-crafted visual odometries \cite{orbslam2, orb-slam3} into the tracking process. Although decoupled methods exhibit better performances in tracking by separating mapping and tracking processes, they require additional computational resources for the independent stages. Moreover, these methods need to store features in 3D space, unrelated to the 3D GS map.

\begin{figure}[t]
  \centering
  \includegraphics[width=8.0cm]{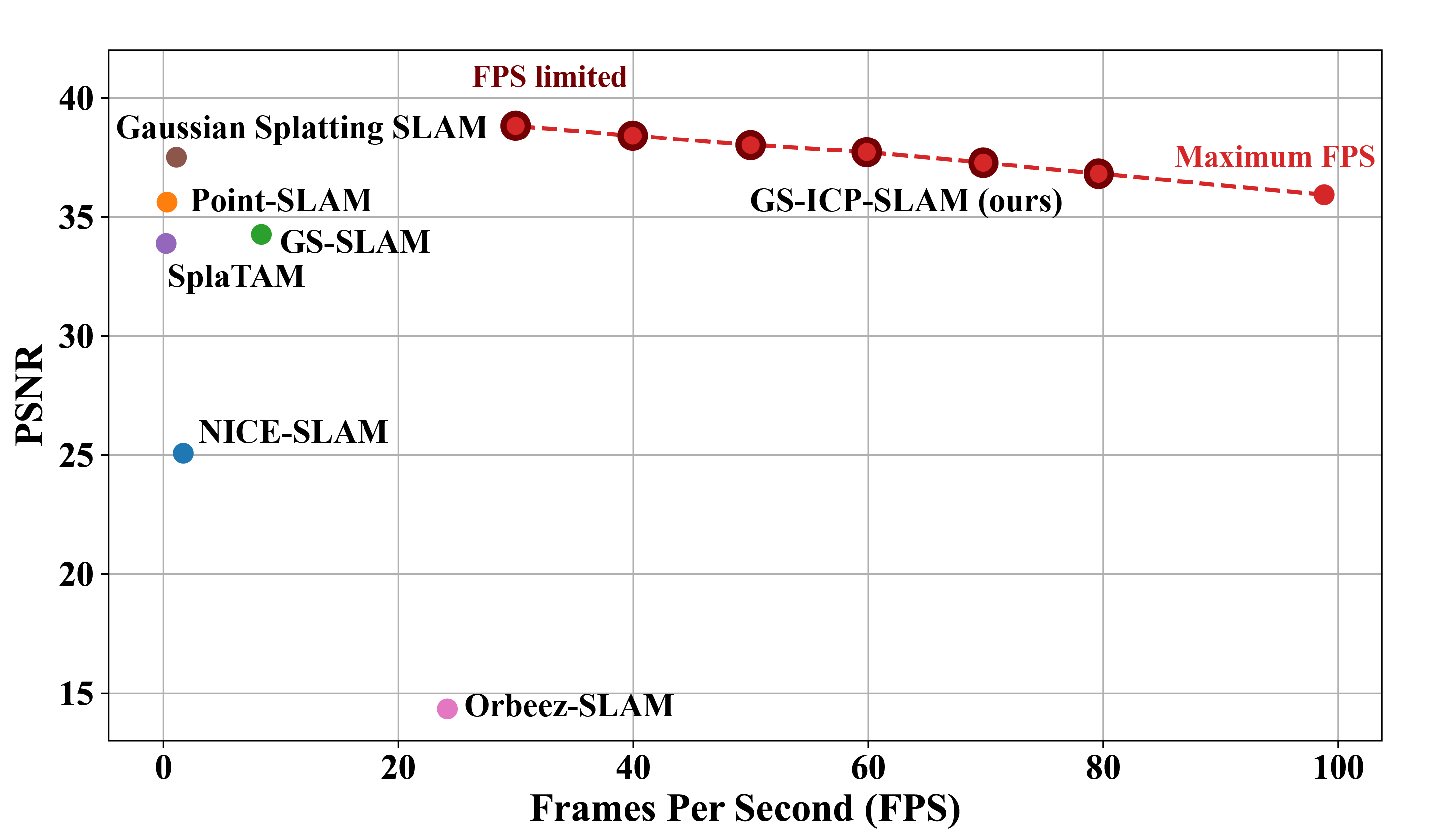}
  \caption{A comparison of PSNR with respect to FPS of entire system in recent research on SLAM algorithm utilizing dense representation such as neural scene representation and 3D Gaussian representation. Our method achieves state-of-the-art performance in rendering evaluation and FPS of entire system. Note that this FPS represents the overall system performance. Reported values are average of Replica 8 scenes.
  }
  \label{fig:title}
   \vspace{-10pt}
\end{figure}

There exists a method that allows Gaussian, an explicit representation, to be directly used for tracking.
The well-known Generalized Iterative Closest Point (G-ICP) \cite{segal2009generalized,koide2021voxelized} from the 3D scan matching family is simple yet efficient for fast tracking of 3D point clouds.
During preprocessing, it only requires computing Gaussians for the current frame and the map.
Given that the map in 3DGS utilizes Gaussians as an explicit representation in 3D space, using G-ICP for tracking allows for the direct utilization of the 3DGS map without the need for post-processing.
Furthermore, the Gaussians of the current frame computed during tracking with G-ICP can also be directly utilized as an explicit representation in the 3DGS map, without additional computations.

Therefore, we propose a dense representation SLAM framework that integrates G-ICP and 3DGS, allowing them to complement each other by sharing explicit representations. 
Unlike traditional methods, our approach actively leverages 3D information through the utilization of G-ICP for tracking. The proposed method is a coupled approach that shares a single map during tracking and mapping processes, while maintaining fast tracking speed akin to decoupled approaches. 
Previous works using decoupled methods \cite{orbeez, kong2023vmap, photoslam} require separate maps containing ORB feature information for tracking, along with additional resources and computations needed to obtain ORB features. Moreover, the values obtained during tracking are not utilized in mapping. 
Our approach exploits the covariance of each point computed during the G-ICP-based tracking process as the initial state of the 3DGS mapping. Meanwhile, the 3D Gaussians in the 3DGS map are also mutually used as 3D points and their covariances which are essential for G-ICP-based tracking, discarding the necessity to recalculate the covariances of the map.
This is possible because the covariances of points computed during the G-ICP process and the 3D Gaussians representing the map commonly contain information about the surrounding space.
In other words, G-ICP and 3DGS can share the same Gaussian world.
Therefore, our system minimizes unnecessary computations and facilitates efficient system configuration by mutually utilizing the key elements, Gaussians, between tracking and mapping processes.
To ensure optimal performance in tracking and mapping sharing information between G-ICP and 3DGS, we also introduce several techniques such as scale alignment. To summarize, our contributions are as follows:
\begin{itemize}
    \item We present a real-time dense representation SLAM that combines G-ICP and 3DGS, achieving extremely high speed of the entire system (up to 107 FPS) and superior quality of the map.
    \item By incorporating G-ICP for tracking, our system utilizes 3D information actively and significantly reduces the time required for the tracking process.
    \item Reduction of computational cost and facilitating rapid convergence of primitives of 3DGS is achieved by sharing the covariances of G-ICP and 3DGS with scale aligning techniques.
\end{itemize}

\section{Related Work}
\textbf{G-ICP} Scan-matching focuses on the registration of two point clouds that observe similar environments by selecting a transformation matrix that minimizes errors between two point clouds \cite{biber2003normal}. Iterative Closest Point (ICP) \cite{icp} algorithm is a widely used and influential method in the scan-matching area. ICP iteratively estimates point correspondences and finds a transformation that minimizes Euclidean distance between corresponding points, thereby optimizing the registration of point clouds. Because of simplicity and speed of ICP, there are many follow-up studies \cite{trimmed-icp, point2planeicp, segal2009generalized, serafin2015nicp, vizzo2023kiss}. 
To improve robustness, Trimmed-ICP \cite{trimmed-icp}, proposed a method for correspondence selection. Point-to-plane ICP \cite{point2planeicp} augmented robustness and accuracy by taking point-to-plane distance as objective function. G-ICP \cite{segal2009generalized} introduced probabilistic models to generalize ICP yielding notable improvements in robustness and accuracy. Scan matching is a common technique used in SLAM for pose tracking \cite{grisetti2010tutorial, palieri2020locus,ceriani2015pose, nieto2007recursive, mallios2014scan}, and for enhanced robustness, feature-based scan matching is often employed \cite{zhang2014loam, shan2018lego, lin2020loam, shan2020lio}.
%
\newline
\textbf{SLAM with Dense Representation}
SLAM with dense representation  \cite{kerl2013dense, czarnowski2020deepfactors} aims to construct maps in a dense form to enable interaction with the map in tasks such as AR, robotics, etc. To achieve this, classic approaches \cite{kinectfusion, dai2017bundlefusion, prisacariu2017infinitam, bylow2013real, whelan2013robust, canelhas2013sdf} creates dense maps by representing space in Signed Distance Field (SDF), and Truncated Signed Distance Field (TSDF), rather than in sparse forms such as point clouds or grids.

Recently, dense representation SLAM methods represent maps by utilizing NeRF \cite{mildenhall2021nerf}, which demonstrates high spatial representation capabilities have been proposed. 
iMAP \cite{imap}, NICE-SLAM \cite{nice-slam}, Point-SLAM \cite{pointslam}, ESLAM \cite{eslam} perform tracking by optimizing camera poses by reducing errors between synthesized and observed images. These methods have limitations in tracking speed and performance. Alternatively, Orbeez-SLAM \cite{orbeez}, vMAP \cite{kong2023vmap} incorporate well-crafted visual SLAM techniques into the tracking process, making mapping and tracking operate independently. Orbeez-SLAM utilizes ORB-SLAM2 \cite{orbslam2} for tracking and instant-ngp \cite{muller2022instant} for learning spatial information to perform mapping. In a similar context, vMAP utilizes ORB-SLAM3 \cite{orb-slam3} for tracking while employing separate Multi-layer perceptrons \cite{taud2018multilayer} for each object to effectively represent the entire space.



3DGS represents space in explicit form using 3D gaussians as primitives. It offers a level of high-fidelity spatial representation similar to NeRF but provides significantly faster rendering speeds. To leverage this advantage to dense representation SLAM, several approaches \cite{gsslam, photoslam, splatam, gaussiansplattingslam} utilizing 3DGS for space representation have been proposed.  
Among them, GS-SLAM \cite{gsslam}, SplaTAM \cite{splatam}, Gaussian Splatting SLAM \cite{gaussiansplattingslam} perform tracking using dense photometric error. While 3DGS shows significantly improved rendering performance compared to NeRF, the tracking methods based on dense photometric error still suffer from tracking speed. To tackle this problem, Photo-SLAM \cite{photoslam} integrates ORB-SLAM3 for tracking, in a similar manner to Orbeez-SLAM and vMAP.



\section{Method}
%
To mutually benefit tracking and mapping, we introduce the fusion of G-ICP and GS, each representing the tracking and mapping processes, respectively. The key insight of our approach is that covariance can be considered as a fundamental common factor for this fusion.
%
Suppose we have a point set (point cloud) $\boldsymbol{\mathcal{X}}=\{\boldsymbol{x}_m\}_{m=1,...,M}$ and its corresponding covariance set
$\boldsymbol{\mathcal{C}}=\{C_m\}_{m=1,...,M}$, where $\boldsymbol{x}=\left[x,y,z\right]^T$.
The covariance $C$ of one 3D point $\boldsymbol{x}$ is given by computing covariance matrix of $k$-nearest neighbors of $\boldsymbol{x}$.
Let us define $\boldsymbol{G}=\{\boldsymbol{\mathcal{X}}, \boldsymbol{\mathcal{C}}\}$ as a set of Gaussian.
G-ICP aims to find a transformation $\mathbf{T}$ that maximally aligns the source Gaussians (current frame) $\boldsymbol{G}^s=\{\boldsymbol{\mathcal{X}}^s, \boldsymbol{\mathcal{C}}^s\}$ and the target Gaussians (map) $\boldsymbol{G}^t=\{\boldsymbol{\mathcal{X}}^t, \boldsymbol{\mathcal{C}}^t\}$.
Assume we know the correspondences between $\boldsymbol{\mathcal{X}}^s$ and $\boldsymbol{\mathcal{X}}^t$ determined by nearest neighbor search.
For example, we have 
$\{\boldsymbol{x}^s_i\}_{i=1,...,N}\subset\boldsymbol{\mathcal{X}}^s$,
$\{C^s_i\}_{i=1,...,N}\subset\boldsymbol{\mathcal{C}}^s$
and 
$\{\boldsymbol{x}^t_i\}_{i=1,...,N}\subset\boldsymbol{\mathcal{X}}^t$,
$\{C^t_i\}_{i=1,...,N}\subset\boldsymbol{\mathcal{C}}^t$,
where $\boldsymbol{x}^s_i$ is associated with $\boldsymbol{x}^t_i$.
To find the optimal transform $\mathbf{T}^*$,
we exploit not the single point but the distribution of that point defined as a Gaussian distribution: $\boldsymbol{x}_i \sim \mathcal{N}(\hat{\boldsymbol{x}}_i, C_i)$. 
Let $d_i = \boldsymbol{x}^t_i - \mathbf{T} \boldsymbol{x}^s_i$ be the error term, and if we assume that there is an optimal transformation $\mathbf{T}^*$, it is clear that $\boldsymbol{x}^t_i = \mathbf{T}^*\boldsymbol{x}^s_i$ thus $\hat{d}_i=\boldsymbol{0}$.
Since we assume $\boldsymbol{x}$ is a Gaussian random variable, $d_i$ is also a Gaussian random variable as the following:
\begin{align}
   \nonumber
   d_i 
   & \sim \mathcal{N} ( \hat{d}_i, C_i^t + \mathbf{T}^* C_i^s (\mathbf{T}^*)^T ) 
   \\
   \nonumber
   &=
   \mathcal{N} ( \hat{x}^t_i - \mathbf{T}^*\hat{x}^s_i, C_i^t + \mathbf{T}^* C_i^s (\mathbf{T}^*)^T )
   \\
   \nonumber
   & =\mathcal{N}(0, C_i^t + \mathbf{T}^* C_i^s (\mathbf{T}^*)^T )
   .
\end{align}
To find the optimal transform $\mathbf{T}^*$ for $\boldsymbol{\mathcal{X}}^s$ and $\boldsymbol{\mathcal{X}}^t$, we use maximum likelihood estimation (MLE) as the following:
\begin{align}    
    \nonumber
    \mathbf{T}^*  
    & = \argmax_\mathbf{T} \prod_i^N p\left(d_i\right)
    = \argmax_\mathbf{T} \sum_i^N \log p(d_i)
    \\
    \label{MLE_for_T}
    & = \argmin_\mathbf{T} \sum_i^N d_i^T \left(C_i^B + \mathbf{T} C_i^A \mathbf{T}^T\right)^{-1} d_i
    .
\end{align}
Therefore, $\mathbf{T}^*$ can be used as the relative pose between the current frame  $\boldsymbol{G}^s=\{\boldsymbol{\mathcal{X}}^s, \boldsymbol{\mathcal{C}}^s\}$ and the map $\boldsymbol{G}^t=\{\boldsymbol{\mathcal{X}}^t, \boldsymbol{\mathcal{C}}^t\}$.

Meanwhile, for mapping purposes, GS also relies on Gaussians $\boldsymbol{G}=\{\boldsymbol{\mathcal{X}}, \boldsymbol{\mathcal{C}}\}$ of the 3D scene representation.
Different from G-ICP, GS aims to find the optimal coordinates of the Gaussians $\boldsymbol{\mathcal{X}}^*=\{\boldsymbol{x}_m^*\}_{m=1,...,M}$ and the optimal covariances $\boldsymbol{\mathcal{C}}^*=\{C_m^*\}_{m=1,...,M}$ as the following:
\begin{align}
    \nonumber
    \label{3DGS_error}
    \boldsymbol{\mathcal{X}}^*, \boldsymbol{\mathcal{C}}^*,
    \boldsymbol{H}^*, \boldsymbol{O}^*
    =
    \argmin_{
    \boldsymbol{\mathcal{X}}, \boldsymbol{\mathcal{C}},
    \boldsymbol{H}, \boldsymbol{O}
    }
    \lambda_{I_1}
    \boldsymbol{\mathcal{L}}_1
    \left(I, I_{gt}\right)
    +
    \lambda_{I_2}
    \boldsymbol{\mathcal{L}}_{D-SSIM}
    \left(I, I_{gt}\right) 
     +
    \lambda_D
    \boldsymbol{\mathcal{L}}_1
    \left(D,D_{gt}\right)
\end{align}
where $\boldsymbol{H}=\{\boldsymbol{h}_m\}_{m=1,...,M}$ and $\boldsymbol{O}=\{\boldsymbol{o}_m\}_{m=1,...,M}$ are the color set and the opacity set of 3D points which are for the RGBD image rendering.
$I$ and $D$ are the rendered RGB and depth images obtained by performing rasterization using $\boldsymbol{G}, \boldsymbol{H}$ and $\boldsymbol{O}$.
%
%
%

\begin{figure}[t]
  \centering
  \includegraphics[scale=0.20]{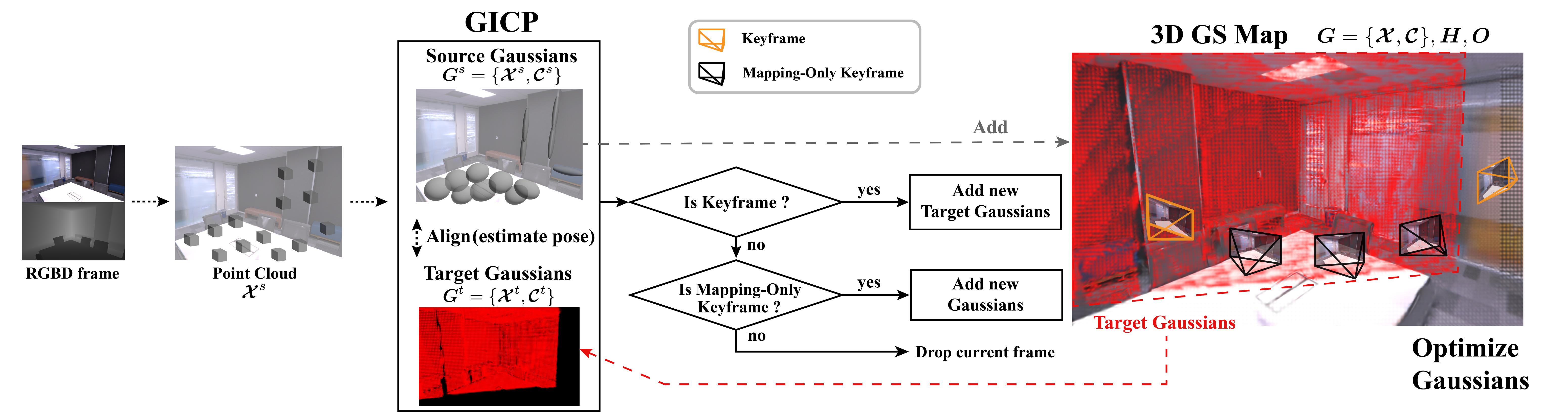}
  \caption{\textbf{System Overview.} The input of our system is RGBD frame. We generate a point cloud by downsampling and reprojecting the current depth image and utilize it in the GICP process. During the GICP process, we create source Gaussians from the point cloud and estimate the current camera pose by aligning them with target Gaussians, which are a subset of the 3D GS map. If the current frame is identified as a keyframe or a mapping-only keyframe, we add the source Gaussians to the 3D GS map as new primitives. Meanwhile, in the mapping process, we optimize the Gaussians along with the color and opacity set of the Gaussians concurrently with the tracking process.
  }
  \label{fig:overview}
   \vspace{-10pt}
\end{figure}
Note that in G-ICP and GS, the key common factor is Gaussians $\boldsymbol{G}=\{\boldsymbol{\mathcal{X}}, \boldsymbol{\mathcal{C}}\}$, allowing these Gaussians to be shared mutually. During G-ICP tracking, the covariance of each frame is computed. Hence, when adding keyframes to expand the 3D GS map, there is no need to recalculate $\boldsymbol{\mathcal{C}}$ for every expansion.
Also, G-ICP does not need to compute the covariances of the map because our GS map already contains Gaussians.
Moreover, G-ICP, by aligning frames based on 3D geometric structure, inherently initializes a certain number of points and their coordinates that aptly represent the 3D structure.
Therefore, it brings about an effect where an appropriate number of points and their poses are initialized, well-suited for depicting the 3D structure, significantly reducing the learning time to find the optimal pose $\boldsymbol{\mathcal{X}}^*$ and the optimal covariance $\boldsymbol{\mathcal{C}}^*$ of Gaussians $\boldsymbol{G}$ in GS.
Additionally, in the same vein, calculations like densifying or opacity reset to adjust the number of 3D points in GS become unnecessary.
Consequently, by sharing a common source, each process becomes mutually beneficial, and the speed of execution accelerates due to reduced redundant computations.
Our method can be simplified as follows:
\begin{enumerate}
    \item Use G-ICP to align the current frame with the 3D GS map which contains covariance (solely need to compute the covariance for the current frame).
    \item When adding keyframes to the 3D GS map, utilize the covariance computed in GICP during tracking (no need for densifying or opacity reset).
\item Repeat steps 1-2.
\end{enumerate}
The overview of our system is shown in Fig.~\ref{fig:overview}.
The detailed implementations of the proposed technique are introduced in the following Section.

\subsection{G-ICP Tracking}
\subsubsection{Scale Regularization} 
When using G-ICP for tracking, aligning the scale of the current frame with the map enables high-performance camera pose estimation \cite{segal2009generalized}.
Given the covariance $C$, scale $\boldsymbol{S}=\left[s_2, s_1, s_0\right]^T$ is given by the singular value decomposition (SVD) as the following:
\begin{align}
  C
  =
  \boldsymbol{R} \boldsymbol{\Lambda}^2 \boldsymbol{R}^T
\end{align}
where $\boldsymbol{\Lambda}=diag\left(s_2, s_1, s_0\right) \forall  s_2>s_1>s_0$ is a scale matrix and
$\boldsymbol{R}$ is the orientation of the Gaussian.
To achieve the robust scan matching performance, point-to-plane ICP or voxelized point-to-plane ICP adopt a regularizing method that makes the scale as $\boldsymbol{S}=\left[1, 1, \epsilon \right]^T$ in order to treat each Gaussian as a plane-like distribution.

In our framework, we utilize target Gaussians $\boldsymbol{G}^t=\{\boldsymbol{\mathcal{X}}^t, \boldsymbol{\mathcal{C}}^t\}$ from the 3D GS map for tracking.
Here, $\boldsymbol{\mathcal{C}}^t$ is optimized to represent the scene accurately, $\boldsymbol{G}^t$ involves not only planes but also lines, corners, and other features.
Therefore, while tracking, instead of regularizing the scale into a plane-based form, preserving the original characteristics of the target Gaussians while performing regularization is more appropriate.
Thus, in our framework, we propose the following ellipse regularization:
\begin{align}
    \boldsymbol{\Lambda}^\prime
    =
    \frac{1}{median\left(\boldsymbol{S}\right)}
    diag \left(s_2, s_1, s_0\right)
\end{align}
where $\boldsymbol{\Lambda}^\prime$ is the ellipse-regularized scale matrix.

\subsubsection{Keyframe Selection}
Similar to \cite{orb-slam, qin2018vins}, we perform dynamic keyframe selection. Considering the geometric structure is critical for both G-ICP tracking and GS mapping, we exploit the geometric correspondence computed from G-ICP.
As Eqn.~\eqref{MLE_for_T} denotes, we can get the distance between $\boldsymbol{x}^s_i$ and $\boldsymbol{x}^t_i$ as an interim result of G-ICP procedure without additional calculations. So we can efficiently determine the correspondence by setting the distance threshold.

\begin{figure*}[t]
    \centering
    \begin{subfigure}[t]{0.5\textwidth}
        \centering
        \includegraphics[scale=0.24]{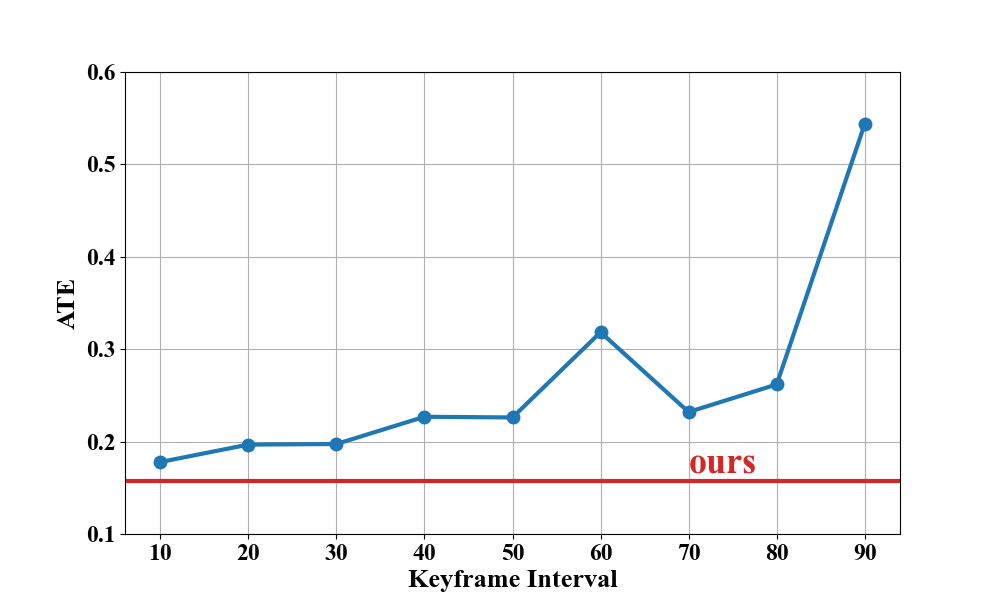}
        \caption{}
         \vspace{-5pt}
    \end{subfigure}%
    ~ 
    \begin{subfigure}[t]{0.5\textwidth}
        \centering
        \includegraphics[scale=0.24]{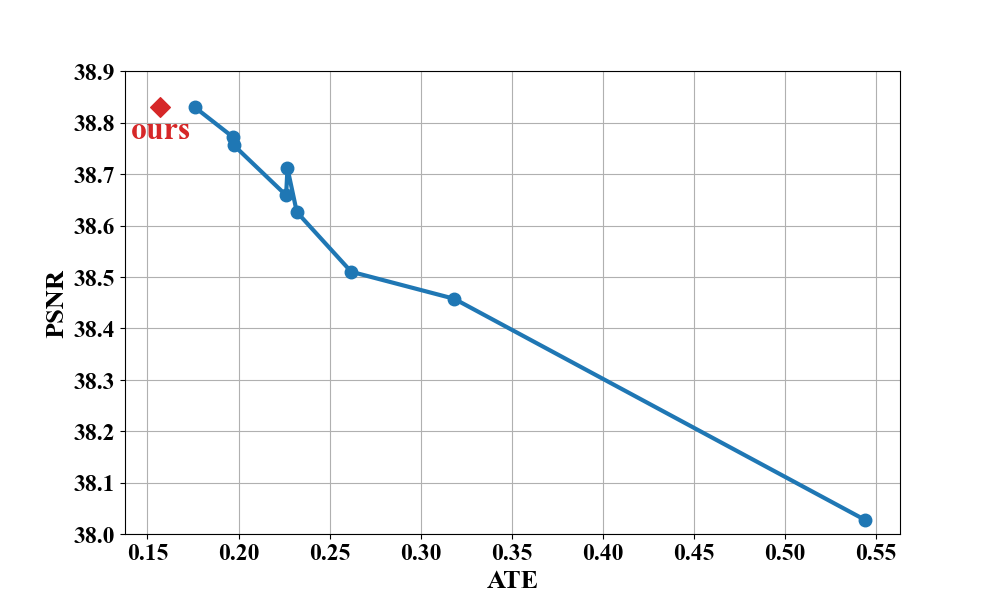}
        \caption{}
         \vspace{-5pt}
    \end{subfigure}
    \caption{\textbf{Tracking Accuracy Comparison Based on Keyframe Selection Methods.} The reported values represent the average results across eight scenes from the Replica dataset \cite{straub2019replica}. When selecting keyframes every n frame (depicted in blue), the tracking accuracy is notably low. Conversely, our keyframe selection method yielded the highest tracking accuracy.}
    \label{fig:keyframe_ate_psnr}
     \vspace{-10pt}
\end{figure*}

By considering the proportion of correspondences between the current frame and map, we select keyframes while taking into account the characteristics of the scene. In more detail, if the proportion in the current frame goes below specific thresholds, the frame is selected as a keyframe. This allows us to achieve consistent tracking performance while simultaneously maintaining a consistent density of Gaussians $\boldsymbol{G}=\{\boldsymbol{\mathcal{X}}, \boldsymbol{\mathcal{C}}\}$ added to the map.
The effect of this keyframe selection is shown in Fig.~\ref{fig:keyframe_ate_psnr} (a).


Note that, from selected keyframe, only Gaussians that do not overlap with the existing current map are considered as target Gaussians. This is because errors due to discrepancies during tracking are conveyed intactly when incorporating into the map. These accumulated errors result in low tracking accuracy. This strategy ensures more accurate tracking results and improves the overall robustness of the system.

\subsection{GS Mapping}

\subsubsection{Scale Aligning} 
Suppose we have a fully trained 3D GS map. Since the 3D GS map is trained to represent the scene from any viewpoint, Gaussians $\boldsymbol{G}$ should be uniformly distributed according to the structure of the 3D space.
However, single frames obtained from radiance sensors such as RGBD cameras or LiDAR yield sparse representations of the 3D space as the distance from the sensor increases, due to the inherent characteristics of the sensor.
In other words, as the distance from the sensor increases, the spacing between 3D points widens, and the scale of the corresponding covariance calculated based on the $k$-nearest neighbors becomes excessively large.
Adding such single frames as keyframes to the map during real-time training of GS mapping leads to an imbalance in the scale of Gaussians, ultimately resulting in a decrease in mapping performance
.
Therefore, to fundamentally alleviate the problem, our proposed method suggests the following scale normalization for the current frame:
\begin{align}
    \nonumber
    \boldsymbol{\Lambda}^{\prime\prime} = \frac{1}{z^p}\boldsymbol{\Lambda}^\prime
    .
\end{align}
Where $\boldsymbol{\Lambda}^{\prime\prime}$ is the normalized scale matrix and $p$ is the parameter which is empirically determined.
By adding such scale-normalized keyframes to the map to expand it, not only does the GS mapping performance increase, but also the performance of G-ICP tracking improves.

\subsubsection{Additional Keyframe Selection for Mapping} 
Mapping based on GS benefits from having a diverse set of frames from various viewpoints for training, as performance improves with a larger number of available frames.
However, increasing the number of selected keyframes during tracking can lead to a degradation in tracking performance due to accumulated errors, as is common with G-ICP-based methods.
To mitigate this issue, we propose adding mapping-only keyframe selection in addition to the existing keyframe selection process. In other words, tracking continues to proceed from selected keyframes in the existing tracking process, while mapping utilizes both the original keyframes and additional mapping-only keyframes.
\begin{figure}[t]
  \centering
  \includegraphics[scale=0.28]{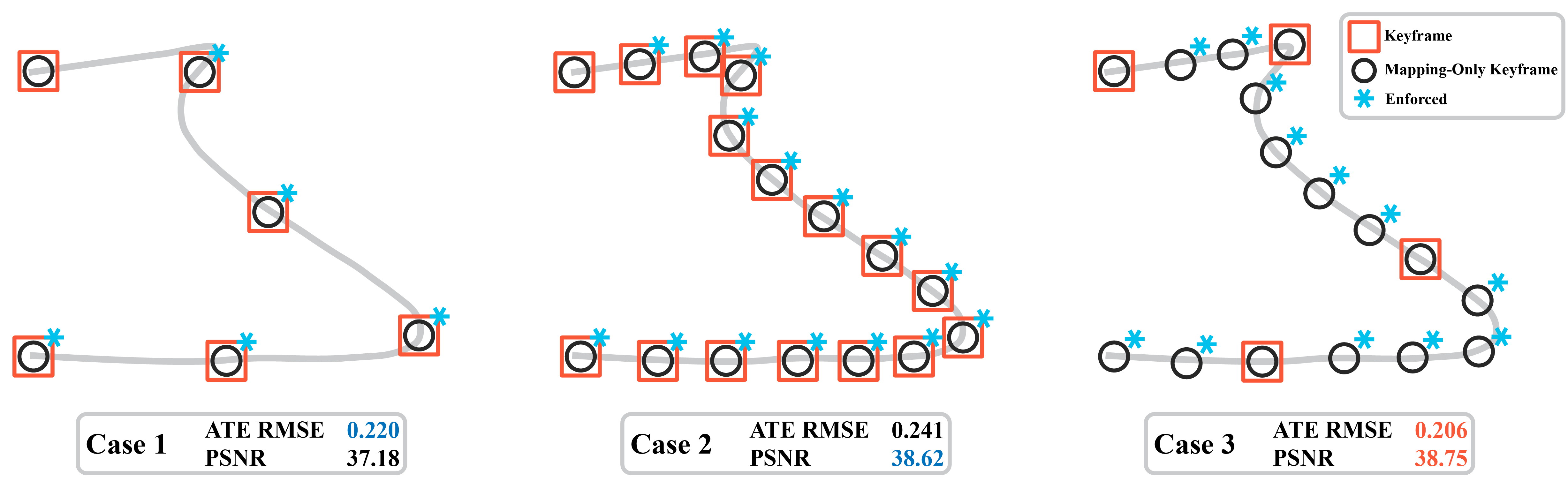}
  \caption{\textbf{Separated Keyframe Selection on Replica office4.} We demonstrate that a small number of tracking keyframes yield accurate trajectory estimation (case 1), while a large number of mapping keyframes result in high rendering performance (case 2). Thus, our method adopts case 3 that select tracking keyframe and mapping keyframe separately at different intervals.
  }
  \label{fig:keyframe}
   \vspace{-10pt}
\end{figure}

The overview of the additional keyframe selection is illustrated in Fig.~\ref{fig:keyframe}.
Cases $1$ and $2$ represent scenarios without considering mapping-only keyframes. In Case $1$, to increase the number of keyframes, if no keyframe satisfying the criteria is found up to the $30^{th}$ frame from the previous keyframe, the $30^{th}$ frame is selected as a keyframe.
However, since this may not yield sufficient mapping performance, as shown in Case $2$, adding an additional keyframe at the $10^{th}$ frame can be considered.
While this approach improves mapping performance as desired, it leads to a decline in tracking performance due to accumulated scan matching errors.
To alleviate this, as shown in Case $3$, we propose adding the $10^{th}$ frame as a mapping-only frame, allowing us to maximize tracking performance while simultaneously maximizing mapping performance.
The effect of the proposed scheme, Case $3$, is shown in Fig.~\ref{fig:keyframe_ate_psnr} (b).
Note that since G-ICP, which performs tracking, computes covariance for all frames anyway, there is no difference in the overall computational load.

\subsubsection{Avoiding Local Minima while Mapping} 
GS SLAM differs from traditional GS in that it operates in real-time and often lacks sufficient observations of the scene. 
Moreover, since it operates in real-time, focusing solely on images from the current viewpoint or nearby viewpoints during training can easily lead to local minima and degrade mapping quality \cite{gaussiansplattingslam}.
For instance, GS simply considers image rendering from specific viewpoints, thus if we continuously train only on a specific viewpoint, the scale of Gaussians in the map tends to elongate in the direction of that viewpoint, leading to local minima.

To overcome this challenge, at each training iteration, we adopt a strategy of randomly choosing one keyframe for learning among the selected keyframes so far, ensuring that both the current observed scene and the entire map are uniformly learned. Additionally, we prune Gaussians that fall into local minima during training to improve mapping performance and ensure robust tracking through geometric preservation.

\section{Experiments}
\subsection{Experimental Setup}
We evaluate the proposed method on Replica dataset \cite{straub2019replica} and TUM dataset \cite{sturm2012benchmark}. Replica dataset contains synthetic scenes and high-quality RGB/depth images rendered from these scenes. TUM dataset includes images captured in the real world, with significant noise and blur, resulting in poor quality. Particularly, many parts of the depth images suffer from information loss. We validate the effectiveness of our approach by evaluating it on both synthetic and real-world datasets. 
All experiments are performed in a desktop with a Ryzen 7 7800x3d CPU, 32GB RAM, and an NVIDIA RTX 4090 24GB GPU.
To evaluate camera tracking accuracy, we use the Root Mean Square Error (RMSE) of the Absolute Trajectory Error (ATE). For the quality of the reconstructed map, we report standard photometric rendering quality metrics (PSNR, SSIM, and LPIPS).
The code will be released upon acceptance of the paper.

\begin{table}[t]
    \centering
    \begin{minipage}{\textwidth} 
    \caption{\textbf{Tracking Performance on Replica (ATE RMSE $\downarrow$ [cm]).} Our method achieves state-of-the-art performance in camera pose estimation. The result of GS-SLAM is taken from \cite{gsslam}. And the result of Photo-SLAM is obtained from \cite{photoslam}.}
    \label{tab:ate_replica}
    \resizebox{9.0cm}{!}{
    \begin{tabular}{l|ccccccccc} 
    \toprule
    Method & R0& R1& R2& Of0& Of1& Of2& Of3& Of4& Avg.\\ 
    \midrule
    NICE-SLAM* \cite{nice-slam}  & 1.61& 1.48& 1.61& 0.95& 0.81& 1.46& 1.76& 1.69 & 1.42   \\
    Point-SLAM* \cite{pointslam} & 0.59 & 0.51 & 0.32 & 0.45 & 0.46 & 0.48 & 0.61 & 0.87 & 0.54 \\
    GS-SLAM \cite{gsslam} & 0.48& 0.53& 0.33& 0.52& 0.41& 0.59& 0.46& 0.70& 0.50\\
    Photo-SLAM \cite{photoslam} & - &-&-&-&-&-&-&-& 0.60 \\
    SplaTAM* \cite{splatam} & 0.29 & 0.35 & 0.28 & 0.49 & 0.21 & 0.31 & 0.34 & 0.57 & 0.36 \\
    \textbf{Ours (limited to 30 FPS)}   & \textbf{0.15} & \textbf{0.16} & \textbf{0.11} & \textbf{0.18} & \textbf{0.12}& \textbf{0.17}& \textbf{0.16}& \textbf{0.21}& \textbf{0.16} \\
    \bottomrule
    \multicolumn{10}{r}{* denotes the reproduced results by running official code.}\\
    \end{tabular}}
    \medskip
%
    \centering
    \caption{\textbf{Tracking Performance on TUM-RGBD} (ATE RMSE $\downarrow$ [cm]). We divide systems into decoupled and coupled systems. Among coupled systems, we achieve competitive performance in camera pose estimation. The result of GS-SLAM is taken from \cite{gsslam}. And the results of ORB-SLAM3 and Photo-SLAM are obtained from \cite{photoslam}.}
    \label{tab:ate_tum}
    \resizebox{9.0cm}{!}{
    \begin{tabular}{l|l|cccc} 
    \toprule
    \multicolumn{2}{c|}{Method} & fr1/desk & fr2/xyz & fr3/office & Avg. \\ 
    \midrule
    \multirow{2}{*}{Decoupled} & ORB-SLAM3 \cite{orb-slam3} & \textbf{1.7} & 0.4 & 1.7 & \textbf{1.3}   \\
    & Photo-SLAM \cite{photoslam} & 2.6 & \textbf{0.3} & \textbf{1.0} & \textbf{1.3} \\
    \midrule
    \multirow{5}{*}{Coupled}
    & NICE-SLAM* \cite{nice-slam} & 2.8 & 2.1 & 7.2 & 4.0 \\
    & Point-SLAM* \cite{pointslam} & \textbf{2.7} & \textbf{1.3} & 3.9 & 2.6 \\
    & GS-SLAM \cite{gsslam} & 3.3 & \textbf{1.3} & 6.6 & 3.7 \\
    & SplaTAM* \cite{splatam} & 3.3 & \textbf{1.3} & 5.1 & 3.2 \\
    & \textbf{Ours (limited to 30 FPS)}   & \textbf{2.7} & 1.8 & \textbf{2.7} & \textbf{2.4} \\ 
    \bottomrule
    \multicolumn{6}{r}{* denotes the reproduced results by running official code.}\\
    \end{tabular}}
    \end{minipage}
    \vspace{-20pt}
\end{table}
\subsection{Camera Tracking Accuracy}
\cref{tab:ate_replica}, \cref{tab:ate_tum} shows the tracking accuracy of our method compared to other approaches on both synthetic \cite{straub2019replica} and real-world \cite{sturm2012benchmark} datasets.
On Replica \cite{straub2019replica}, the proposed method achieved state-of-the-art (SOTA) performance across all scenes, reducing the trajectory error by more than 50\% compared to the previous SOTA. This result arises from the fact that our proposed method actively utilizes 3D information by employing G-ICP \cite{segal2009generalized} for tracking, unlike other methods that perform tracking based on errors in 2D space. On TUM \cite{sturm2012benchmark}, our method outperforms coupled methods \cite{nice-slam, pointslam, gsslam, splatam} which conduct tracking and mapping based on a single map. 
Decoupled method \cite{photoslam} and its baseline method \cite{orb-slam3} show better performance, but these approaches require additional resources for storing separate map information and costly computations for extracting features solely for the tracking process.

\begin{table}[hb!]
\vspace{-10pt}
\caption{\textbf{Evaluation of System Speed and Map Quality on Replica.} Our method outperforms all other frameworks in both system speed and quality of the reconstructed map.}
\label{tab:rendering_replica}
\centering\resizebox{\textwidth}{!}{
\begin{tabular}{l|c|cccccccccc}
\toprule
Methods                                      & Metrics                 & R0            & R1             & R2             & Of0           
                                                                     & Of1          & Of2           & Of3           & Of4           & Avg.          \\
\midrule
                                                                     
                                                                     


\multirow{4}{*}{Orbeez-SLAM* \cite{orbeez}}                          
                                            & PSNR[dB] $\uparrow$    & 12.13       & 15.28        & 15.87        & 17.59        
                                                                     & 19.26       & 10.30        & 11.55        & 12.65        & 14.33    \\
                                                                     
                                            & SSIM $\uparrow$        & 0.699       & 0.787        & 0.808        & 0.795        
                                                                     & 0.672       & 0.783        & 0.760        & 0.838        & 0.768    \\
                                                                     
                                            & LPIPS $\downarrow$     & 0.612       & 0.477        & 0.509        & 0.515        
                                                                     & 0.413       & 0.554        & 0.540        & 0.528        & 0.529    \\

                                            & FPS $\uparrow$         & 24.05            & 24.78             & 24.00             & 19.22             
                                                                     & 26.74            & 23.88             & 24.45             & 26.11             & 24.15         \\

\midrule
\multirow{4}{*}{Point-SLAM* \cite{pointslam}}
                                            & PSNR[dB] $\uparrow$    & 33.38            & 34.10             & 36.32             & 38.72                  
                                                                     & 39.31            & 34.22             & 34.10             & 34.82             & 35.62         \\

                                            & SSIM $\uparrow$        & \textbf{0.979}   & \textbf{0.977}    & \textbf{0.985}    & 0.985                  
                                                                     & 0.987            & 0.962             & 0.963             & \textbf{0.981}    & \textbf{0.977} \\
                                                                     
                                            & LPIPS $\downarrow$     & 0.097            & 0.115             & 0.101             & 0.089                  
                                                                     & 0.110            & 0.152             & 0.119             & 0.131             & 0.114         \\        
                                            
                                            & FPS $\uparrow$         & 0.26            & 0.30             & 0.31             & 0.33                  
                                                                     & 0.34            & 0.30             & 0.28             & 0.30             & 0.30         \\
\midrule

\multirow{4}{*}{GS-SLAM \cite{gsslam}}
                                            & PSNR[dB] $\uparrow$    & 31.56            & 32.86             & 32.59             &  38.70            
                                                                     &  41.17           &  32.36            &  32.03            &  32.92            & 34.27         \\
                                                                     
                                            & SSIM $\uparrow$        &  0.968           & 0.973             & 0.971             &  \textbf{0.986}            
                                                                     &  \textbf{0.993}  & \textbf{0.978}    & \textbf{0.970}   &  0.968            & 0.975         \\
                                                                     
                                            & LPIPS $\downarrow$     & 0.094            & 0.075             & 0.093             &  0.050            
                                                                     &  0.033           &  0.094            &  0.110            &  0.112            & 0.082         \\

                                            & FPS $\uparrow$         & 8.34          & -                  & -                  & -                  
                                                                     & -                 & -                  & -                  & -                  & 8.34
                                                                     \\


                                                                     
                                            
\midrule

\multirow{4}{*}{SplaTAM* \cite{splatam}}
                                            & PSNR[dB] $\uparrow$    & 32.60            & 33.55             & 34.83             & 38.09             
                                                                     & 39.02            & 31.95             & 29.53             & 31.55             & 33.89    \\

                                            & SSIM $\uparrow$        & 0.975            & 0.969             & 0.982             & 0.982              
                                                                     & 0.982            & 0.966             & 0.949             & 0.951             & 0.970     \\
                                                                     
                                            & LPIPS $\downarrow$     & 0.070            & 0.097             & 0.074             & 0.088              
                                                                     & 0.093            & 0.098             & 0.119             & 0.150             & 0.099     \\        
                                            
                                            & FPS $\uparrow$         & 0.24            & 0.19             & 0.19             & 0.20                  
                                                                     & 0.22            & 0.27             & 0.26             & 0.24             & 0.23     \\
\midrule


                                                                     
                                            

\multirow{4}{*}{\textbf{Ours (no tracking speed limit)}}
                                            & PSNR[dB] $\uparrow$    & 32.20            & 35.36             & 34.42             & 40.31        
                                                                     & 40.75            & 33.85             & 34.08             & 36.47             & 35.93    \\
                                                                     
                                            & SSIM $\uparrow$        & 0.940            & 0.960             & 0.957             & 0.978        
                                                                     & 0.977            & 0.962             & 0.953             & 0.963             & 0.962    \\
                                            
                                            & LPIPS $\downarrow$     & 0.081            & 0.067             & 0.083             & 0.045        
                                                                     & 0.051            & 0.069             & 0.067             & 0.065             & 0.066    \\

                                            & FPS $\uparrow$         & \textbf{100.98}           & \textbf{84.92}        & \textbf{103.38}        & \textbf{99.10}        
                                                                     & \textbf{107.06}           & \textbf{95.60}        & \textbf{97.20}         & \textbf{96.66}        & \textbf{98.11}     \\
\midrule

\multirow{4}{*}{\textbf{Ours (limited to 30 FPS)}}                                                                  
                                            & PSNR[dB] $\uparrow$    & \textbf{35.37}   & \textbf{37.80}    & \textbf{38.50}        & \textbf{43.13}        
                                                                     & \textbf{43.26}   & \textbf{36.93}    & \textbf{36.90}        & \textbf{38.75}        & \textbf{38.83}    \\
                                                                     
                                            & SSIM $\uparrow$        & 0.963            & 0.971        & 0.975        & \textbf{0.986}        
                                                                     & 0.985            & 0.974        & 0.969        & 0.973        & 0.975    \\
                                            
                                            & LPIPS $\downarrow$     & \textbf{0.048}            & \textbf{0.045}        & \textbf{0.048}        & \textbf{0.026}        
                                                                     & \textbf{0.029}            & \textbf{0.043}        & \textbf{0.042}        & \textbf{0.045}        & \textbf{0.041}    \\

                                            & FPS $\uparrow$         & 29.97            & 29.98        & 29.98        & 29.98        
                                                                     & 29.99            & 29.97        & 29.97        & 29.97        & 29.98    \\
\bottomrule
\multicolumn{11}{r}{* denotes the reproduced results by running official code.}\\
\multicolumn{11}{r}{
The result of GS-SLAM is taken from \cite{gsslam}.
}
\vspace{-10pt}
\end{tabular}}
\end{table}

\begin{table}[hbt!]
\caption{\textbf{Evaluation of System Speed and Map Quality on TUM-RGBD.} Proposed method shows incredible system speed and competitive map quality.}
\label{tab:rendering_tum}
\centering\resizebox{8.0cm}{!}{
\begin{tabular}{l|cccc}
\toprule
Methods                 
                        & PSNR[dB] $\uparrow$  & SSIM $\uparrow$  & LPIPS $\downarrow$    & FPS $\uparrow$   \\
\midrule
{NICE-SLAM* \cite{nice-slam}}                                                                       
                        & 14.10            & 0.574             & 0.395     & 0.08       \\
\midrule

{Point-SLAM* \cite{pointslam}}
                        & 21.40            & 0.738             & 0.447  & 0.22     \\

\midrule

{Photo-SLAM \cite{photoslam}}
                  & 21.90           & 0.763            & 0.187    & -    \\

\midrule

{SplaTAM* \cite{splatam}}
                      & \textbf{23.46}            & \textbf{0.906}             & \textbf{0.156}        & 0.32     \\

\midrule

{\textbf{Ours (unlimited tracking speed)}}
                   & 19.62            & 0.750             & 0.240        & \textbf{73.92}      \\

\midrule

{\textbf{Ours (limited to 30 FPS)}}                                                                  
                  & 20.72            & 0.768             & 0.218          & 29.99     \\

\bottomrule
\multicolumn{5}{r}{* denotes the reproduced results by running official code.}\\
\multicolumn{5}{r}{The result of Photo-SLAM is taken from \cite{photoslam}.}
\end{tabular}}
\end{table}
\subsection{System Speed and Quality of Reconstructed Map}
\cref{tab:rendering_replica} and \cref{tab:rendering_tum} demonstrates the FPS of the systems and quality of the reconstructed map on Replica \cite{straub2019replica} and TUM \cite{sturm2012benchmark} dataset, respectively. FPS of the system is calculated by dividing the total number of frames by the total time. Note that this FPS represents the overall system performance, encompassing the entire system including tracking and mapping processes, rather than just individual components. Photo-SLAM \cite{photoslam} reported the tracking FPS of their system, but did not include the system FPS, so we leave that field blank in the \cref{tab:rendering_tum}.

Since our method implements mapping and tracking processes to operate in parallel, the number of map optimization iterations varies depending on the tracking speed, consequently affecting the quality of the reconstructed map. So we evaluate our system in two cases: one where the tracking speed is limited to the typical sensor input speed of 30 FPS, and the other where the tracking speed is not limited.

\begin{figure}[t]
  \centering
  \includegraphics[width=12.0cm]{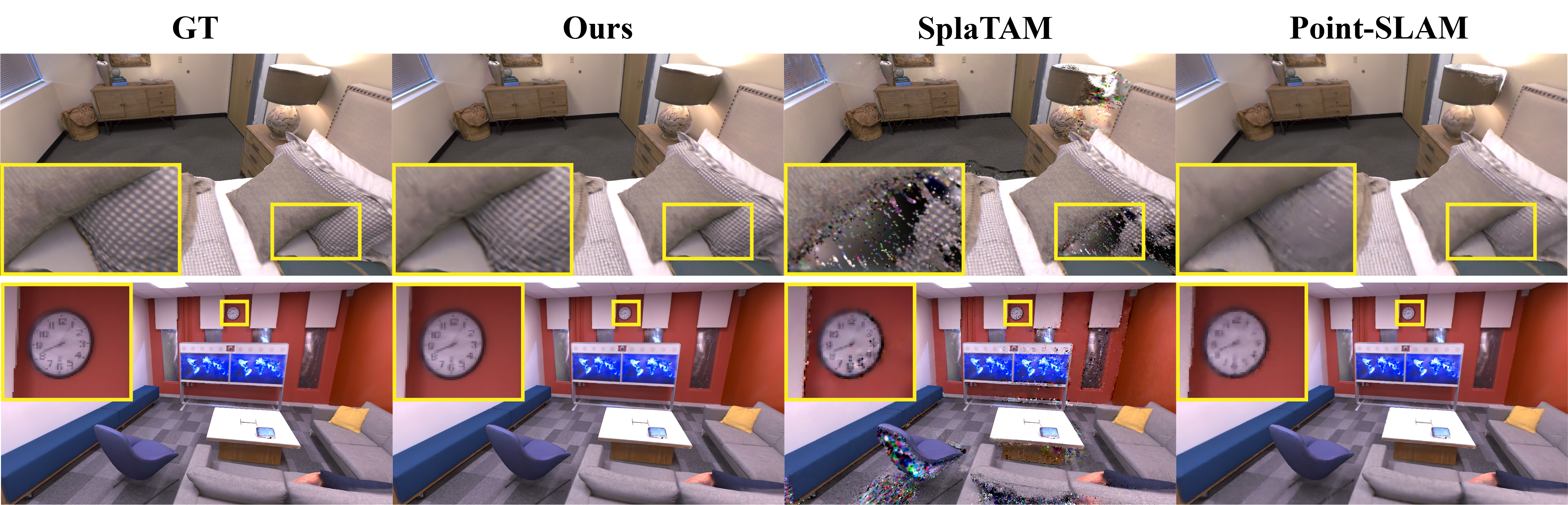}
  \caption{\textbf{Comparison of Rendering Results.} In the first scene, SplaTAM \cite{splatam} failed to reconstruct the pillow and lamp. Point-SLAM \cite{pointslam} failed to represent the detailed pattern of the pillow. In the second case, SplaTAM and Point-SLAM failed to accurately reconstruct the details of the clock. However, our method exhibits rendering results that closely resemble the ground truth image, demonstrating a high level of accuracy.}
  \label{fig:rendering_comparison}
   \vspace{-10pt}
\end{figure}

In the limited case (30 FPS), our system achieves SOTA map quality across all scenes in the Replica dataset. When the tracking speed is not limited, our method shows extremely fast system speed up to 107 FPS. This speed is more than four times faster than that of the Orbeez-SLAM \cite{orbeez}, based on the ORB-SLAM2 \cite{orbslam2}. Remarkably, even under this condition, our approach maintains superior map quality and outperforms other methods. The proposed method effectively integrates Gaussians for map representation with appropriate initial states and has better-detailed representation capability compared to NeRF-based methods. As shown in \cref{fig:rendering_comparison}, SplaTAM suffers from incorporating appropriate Gaussians into the map, leading to regions of poor quality. Point-SLAM exhibits areas where detailed representation is lacking.

In TUM-RGBD, our method demonstrates remarkably fast system speed with competitive map quality, showcasing its practicality in real-world scenarios. Compared to SplaTAM \cite{splatam}, our approach shows a slight decrease in map quality, with a PSNR reduction of approximately 11.7\%. However, we achieve significant speed improvements, with a speedup of approximately 91.6 times under limited conditions and up to 227 times at maximum speed. The reason is that, unlike Replica, TUM dataset is captured with old-fashioned sensors, resulting in noisy and substantial information loss in depth images. While SplaTAM treats these factors by adding new Gaussians only in regions with significant depth loss, our method utilizes all structural information from the depth image, focusing on accurate tracking and rapid system rather than treating such factors.

\subsection{Ablation study}

\begin{table}[h]
\vspace{-10pt}
\caption{\textbf{Scale Regularization Ablation on TUM-RGBD.} Reported results are the average ATE RMSE $\downarrow$ [cm] of 3 scenes in TUM-RGBD dataset.}
\label{tab:regularization}
\centering{
\begin{tabular}{c|cccc}
\toprule
Scale Regularization   & fr1-desk   & fr2-xyz  & fr3-office & Avg. \\
\midrule
\xmark & 136.32 & 196.61 & 376.40 & 236.54 \\
Plane  & 65.63  & 5.53   & 16.21  & 29.12   \\
Ellipse & \textbf{2.66}  & \textbf{1.77}  & \textbf{2.67}  & \textbf{2.37} \\
\bottomrule
\end{tabular}}
\vspace{-5pt}
\end{table}

\noindent\textbf{Scale Regularization}
\cref{tab:regularization} shows the result of the ablation study on scale regularization while tracking. 
We perform scale regularization to enhance tracking accuracy by ensuring that the scales of Gaussians in the current frame and the map are within a similar range. Despite the application of plane regularization resulting in some improvement in tracking accuracy, it remains inadequate. This is because the plane regularization method considers all the Gaussians as planes, ignoring the characteristics of Gaussians existing in the map, which are optimized to represent the surrounding space through the mapping process. On the other hand, when applying the proposed ellipse regularization, the best tracking accuracy is achieved. This is attributed to the ability of the proposed ellipse regularization to consider the characteristics of Gaussians existing in the map while performing regularization. 
\begin{table}[ht]
\vspace{-5pt}
\caption{\textbf{Ablation of Scale Aligning on Replica.} The results are the average of 8 scenes in the Replica dataset.}
\label{tab:scale_re_init_ablation}
\centering\resizebox{0.6\textwidth}{!}{
\begin{tabular}{cc|cccc}
\toprule
Covs from G-ICP & Re-init  & ATE $\downarrow$  & PSNR $\uparrow$      & SSIM $\uparrow$      & LPIPS $\downarrow$     \\
\midrule
\xmark      & \xmark                                        & 8.893     & 24.81     & 0.840     & 0.380     \\
\checkmark  & \xmark                                        & 0.258     & 33.21     & 0.922     & 0.177     \\
\checkmark  & \textit{constant}                                & 0.158     & 37.33     & 0.964     & 0.074     \\
\checkmark  & $z$                             & 0.158     & 38.70     & 0.974     & 0.044     \\
\checkmark  & $z^{1.5}$                       & \textbf{0.157}     & \textbf{38.83}     & \textbf{0.975}     & \textbf{0.041}     \\
\bottomrule
\end{tabular}}
\vspace{-10pt}
\end{table}
\newline
\noindent\textbf{Scale Aligning}
We implement scale aligning to reduce the difference between scales of existing Gaussians in the map and newly added Gaussians. By doing this, we ensure that scales of newly added Gaussians closely resemble the optimal value they will reach through optimization. To demonstrate the effectiveness of this approach, we tested our method in various cases and \cref{tab:scale_re_init_ablation} shows the results. 
When not utilizing the covariances computed during the G-ICP \cite{segal2009generalized} process, we calculate the scales of Gaussians by using simple-knn module of vanilla 3DGS \cite{gaussiansplatting}, and set rotations as identity. In this case, a significant drop in tracking performance appeared. This is because the newly added Gaussians lack sufficient prior information about the 3D structure, causing them to be utilized as target Gaussians without being optimally optimized in space.
On the other hand, in the case of using the computed covariance from the G-ICP process, the tracking and mapping performance improves. Furthermore, the best performance is observed when dividing the scale by $z^{1.5}$.
These results indicate that utilizing covariances calculated during G-ICP procedure is effective for both tracking and mapping, and scale aligning facilitates smooth connection from G-ICP to 3DGS. 
\begin{table}[ht]
\vspace{-5pt}
\caption{\textbf{Keyframe and Mapping-Only Keyframe Selection Ablation on Replica.} Reported results are the average of 8 scenes in the Replica dataset.}
\label{tab:keyframe_separation}
\centering\resizebox{0.85\textwidth}{!}{
\begin{tabular}{l|l|cccc}
\toprule
Keyframe & Mapping-Only Keyframe    & ATE [cm] $\downarrow$  & PSNR $\uparrow$      & SSIM $\uparrow$      & LPIPS $\downarrow$ \\
\midrule
tracking + every 30 frame & \multicolumn{1}{c|}{-} & 0.174 & 37.52 & 0.968 & 0.055 \\
tracking + every 10 frame & \multicolumn{1}{c|}{-} & 0.176 & \textbf{38.83} & 0.974 & \textbf{0.041} \\
\midrule
tracking & every 10 frame & \textbf{0.157} & \textbf{38.83} & \textbf{0.975} & \textbf{0.041} \\
\bottomrule
\end{tabular}}
\vspace{-10pt}
\end{table}
\newline
\noindent\textbf{Additional Keyframe Selection for Mapping}
For mapping quality, sufficient keyframes are required, but indiscriminate addition of keyframes causes a drop in tracking accuracy. To address this challenge and maximize both tracking accuracy and mapping quality, we pick mapping-only keyframes in addition to the original keyframes. \cref{tab:keyframe_separation} presents the result of the ablation study of this method. When using only keyframes without mapping-only keyframes, camera tracking accuracy suffers due to accumulated errors in scan-matching. However, employing our proposed method yields the best tracking accuracy and mapping quality. This result shows that our method selectively adds keyframes essential for tracking, thereby minimizing accumulated errors in scan-matching. While leveraging these benefits, we ensure sufficient primitives and RGBD images for training, which are crucial for enhancing map quality through the use of mapping-only keyframes.
\begin{table}[ht]
\caption{\textbf{Ablation of Methods for Avoiding Local Minima While Mapping on Replica.} The results are the average of 8 scenes in the Replica dataset.}
\vspace{-5pt}
\label{tab:ablation_local_minima}
\resizebox{1.07\textwidth}{!}{
\begin{subtable}{0.45\textwidth}
    \label{tab:train_keyframe}
    \begin{tabular}{c|ccc}
    \multicolumn{4}{c}{(a)} \\
    \toprule
    Keyframe choice for learning  & PSNR      & SSIM      & LPIPS  \\
    \midrule
    recent 1 keyframe     & 26.85 & 0.872 & 0.200 \\
    covisible keyframes   & 31.34 & 0.924 & 0.120 \\
    random keyframes      & \textbf{38.83} & \textbf{0.975} & \textbf{0.041} \\
    \bottomrule
    \end{tabular}
\end{subtable}
~
\begin{subtable}{0.7\textwidth}
    \label{tab:managing_gaussians}
    \centering{}{
    \begin{tabular}{cc|cccc}
    \multicolumn{6}{c}{(b)} \\
    \toprule
    densifying  & pruning       & ATE [cm]  & PSNR       & SSIM       & LPIPS \\
    \midrule
    \checkmark  & \checkmark    & 0.188 & 38.65 & 0.973 & 0.041 \\
    \checkmark  & \xmark        & 0.170 & 37.90 & 0.971 & 0.052 \\
    \xmark      & \checkmark    & \textbf{0.157} & \textbf{38.83} & \textbf{0.975} & \textbf{0.041} \\
    \xmark      & \xmark        & 0.159 & 38.72 & 0.974 & 0.041 \\
    \bottomrule
    \end{tabular}}
\end{subtable}}
\vspace{-3pt}
\end{table}
\newline
\noindent\textbf{Avoiding Local Minima while Mapping}
We employ two methods to prevent Gaussians from overfitting to the training image, which results in elongation to the viewpoint direction: (1) training with randomly selected keyframes and (2) Gaussian prunning. \cref{tab:ablation_local_minima} presents the results of the ablation study on these methods. 

In \cref{tab:train_keyframe}, the worst rendering performance is observed when repeatedly training with the most recently added keyframe, while using covisible keyframes for training yields better results. Covisible keyframes offer more diverse viewpoints for the Gaussians compared to a single keyframe, yet they still provide limited diversity in viewpoints. Therefore, utilizing them for training may still result in reduced map quality. Random keyframes provide the most diverse viewpoints compared to other methods, leading to a significant improvement in map quality.

In the vanilla 3DGS \cite{gaussiansplatting}, Gaussian densifying/pruning techniques are utilized to manage Gaussians. \cref{tab:managing_gaussians} shows the results of an ablation test of Gaussian densifying/pruning in our method. Since our method appropriately supplies Gaussians for spatial representation, densifying is unnecessary. However, pruning improves tracking accuracy and map quality by removing Gaussians that have become elongated due to overfitting or are no longer essential for representing the map.

\section{Conclusion}
In this paper, we have proposed RGBD GS-ICP-SLAM, a dense representation SLAM system that leverages 3D Gaussian representation for high-fidelity spatial representation. We demonstrate that a fusion of G-ICP and 3DGS that utilizes a single 3D Gaussian map for both tracking and mapping yields mutual benefits. The exchange of Gaussians between tracking and mapping processes with scale alignments minimizes redundant computations and constructs an efficient system. Moreover, our dynamic keyframe selection method enhances both tracking and mapping performance. Through extensive experiments, the proposed approach presents state-of-the-art performance in spatial representation, camera pose estimation, and total system speed.
%
\newline
\textbf{Limitations.}
The proposed method achieved fast system speed by relying solely on depth for the 3D structure. However, in real-world environments, there are limitations in the quality of reconstructed maps due to the inherent depth noise in RGB-D sensors. Since the system speed is already exceptionally fast, it is expected that robust performance can be achieved in noisy real-world environments by trading off a bit of speed to compensate for noisy depth images with relatively robust RGB information.




%
%
{
\bibliographystyle{splncs04}
\bibliography{main}

\begin{thebibliography}{10}
\providecommand{\url}[1]{\texttt{#1}}
\providecommand{\urlprefix}{URL }
\providecommand{\doi}[1]{https://doi.org/#1}

\bibitem{icp}
Besl, P.J., McKay, N.D.: Method for registration of 3-d shapes. In: Sensor fusion IV: control paradigms and data structures. vol.~1611, pp. 586--606. Spie (1992)

\bibitem{biber2003normal}
Biber, P., Stra{\ss}er, W.: The normal distributions transform: A new approach to laser scan matching. In: Proceedings 2003 IEEE/RSJ International Conference on Intelligent Robots and Systems (IROS 2003)(Cat. No. 03CH37453). vol.~3, pp. 2743--2748. IEEE (2003)

\bibitem{bylow2013real}
Bylow, E., Sturm, J., Kerl, C., Kahl, F., Cremers, D.: Real-time camera tracking and 3d reconstruction using signed distance functions. In: Robotics: Science and Systems. vol.~2, p.~2 (2013)

\bibitem{orb-slam3}
Campos, C., Elvira, R., Rodr{\'\i}guez, J.J.G., Montiel, J.M., Tard{\'o}s, J.D.: Orb-slam3: An accurate open-source library for visual, visual--inertial, and multimap slam. IEEE Transactions on Robotics  \textbf{37}(6),  1874--1890 (2021)

\bibitem{canelhas2013sdf}
Canelhas, D.R., Stoyanov, T., Lilienthal, A.J.: Sdf tracker: A parallel algorithm for on-line pose estimation and scene reconstruction from depth images. In: 2013 IEEE/RSJ International Conference on Intelligent Robots and Systems. pp. 3671--3676. IEEE (2013)

\bibitem{ceriani2015pose}
Ceriani, S., S{\'a}nchez, C., Taddei, P., Wolfart, E., Sequeira, V.: Pose interpolation slam for large maps using moving 3d sensors. In: 2015 IEEE/RSJ international conference on intelligent robots and systems (IROS). pp. 750--757. IEEE (2015)

\bibitem{trimmed-icp}
Chetverikov, D., Svirko, D., Stepanov, D., Krsek, P.: The trimmed iterative closest point algorithm. In: 2002 International Conference on Pattern Recognition. vol.~3, pp. 545--548. IEEE (2002)

\bibitem{orbeez}
Chung, C.M., Tseng, Y.C., Hsu, Y.C., Shi, X.Q., Hua, Y.H., Yeh, J.F., Chen, W.C., Chen, Y.T., Hsu, W.H.: Orbeez-slam: A real-time monocular visual slam with orb features and nerf-realized mapping. In: 2023 IEEE International Conference on Robotics and Automation (ICRA). pp. 9400--9406. IEEE (2023)

\bibitem{czarnowski2020deepfactors}
Czarnowski, J., Laidlow, T., Clark, R., Davison, A.J.: Deepfactors: Real-time probabilistic dense monocular slam. IEEE Robotics and Automation Letters  \textbf{5}(2),  721--728 (2020)

\bibitem{dai2017bundlefusion}
Dai, A., Nie{\ss}ner, M., Zollh{\"o}fer, M., Izadi, S., Theobalt, C.: Bundlefusion: Real-time globally consistent 3d reconstruction using on-the-fly surface reintegration. ACM Transactions on Graphics (ToG)  \textbf{36}(4), ~1 (2017)

\bibitem{grisetti2010tutorial}
Grisetti, G., K{\"u}mmerle, R., Stachniss, C., Burgard, W.: A tutorial on graph-based slam. IEEE Intelligent Transportation Systems Magazine  \textbf{2}(4),  31--43 (2010)

\bibitem{photoslam}
Huang, H., Li, L., Cheng, H., Yeung, S.K.: Photo-slam: Real-time simultaneous localization and photorealistic mapping for monocular, stereo, and rgb-d cameras. arXiv preprint arXiv:2311.16728  (2023)

\bibitem{eslam}
Johari, M.M., Carta, C., Fleuret, F.: Eslam: Efficient dense slam system based on hybrid representation of signed distance fields. In: Proceedings of the IEEE/CVF Conference on Computer Vision and Pattern Recognition. pp. 17408--17419 (2023)

\bibitem{splatam}
Keetha, N., Karhade, J., Jatavallabhula, K.M., Yang, G., Scherer, S., Ramanan, D., Luiten, J.: Splatam: Splat, track \& map 3d gaussians for dense rgb-d slam. arXiv preprint arXiv:2312.02126  (2023)

\bibitem{gaussiansplatting}
Kerbl, B., Kopanas, G., Leimk{\"u}hler, T., Drettakis, G.: 3d gaussian splatting for real-time radiance field rendering. ACM Transactions on Graphics  \textbf{42}(4) (2023)

\bibitem{kerl2013dense}
Kerl, C., Sturm, J., Cremers, D.: Dense visual slam for rgb-d cameras. In: 2013 IEEE/RSJ International Conference on Intelligent Robots and Systems. pp. 2100--2106. IEEE (2013)

\bibitem{koide2021voxelized}
Koide, K., Yokozuka, M., Oishi, S., Banno, A.: Voxelized gicp for fast and accurate 3d point cloud registration. In: 2021 IEEE International Conference on Robotics and Automation (ICRA). pp. 11054--11059. IEEE (2021)

\bibitem{kong2023vmap}
Kong, X., Liu, S., Taher, M., Davison, A.J.: vmap: Vectorised object mapping for neural field slam. In: Proceedings of the IEEE/CVF Conference on Computer Vision and Pattern Recognition. pp. 952--961 (2023)

\bibitem{lin2020loam}
Lin, J., Zhang, F.: Loam livox: A fast, robust, high-precision lidar odometry and mapping package for lidars of small fov. In: 2020 IEEE International Conference on Robotics and Automation (ICRA). pp. 3126--3131. IEEE (2020)

\bibitem{point2planeicp}
Low, K.L.: Linear least-squares optimization for point-to-plane icp surface registration. Chapel Hill, University of North Carolina  \textbf{4}(10), ~1--3 (2004)

\bibitem{mallios2014scan}
Mallios, A., Ridao, P., Ribas, D., Hern{\'a}ndez, E.: Scan matching slam in underwater environments. Autonomous Robots  \textbf{36},  181--198 (2014)

\bibitem{gaussiansplattingslam}
Matsuki, H., Murai, R., Kelly, P.H., Davison, A.J.: Gaussian splatting slam. arXiv preprint arXiv:2312.06741  (2023)

\bibitem{mildenhall2021nerf}
Mildenhall, B., Srinivasan, P.P., Tancik, M., Barron, J.T., Ramamoorthi, R., Ng, R.: Nerf: Representing scenes as neural radiance fields for view synthesis. Communications of the ACM  \textbf{65}(1),  99--106 (2021)

\bibitem{muller2022instant}
M{\"u}ller, T., Evans, A., Schied, C., Keller, A.: Instant neural graphics primitives with a multiresolution hash encoding. ACM Transactions on Graphics (ToG)  \textbf{41}(4),  1--15 (2022)

\bibitem{orb-slam}
Mur-Artal, R., Montiel, J.M.M., Tardos, J.D.: Orb-slam: a versatile and accurate monocular slam system. IEEE transactions on robotics  \textbf{31}(5),  1147--1163 (2015)

\bibitem{orbslam2}
Mur-Artal, R., Tard{\'o}s, J.D.: Orb-slam2: An open-source slam system for monocular, stereo, and rgb-d cameras. IEEE transactions on robotics  \textbf{33}(5),  1255--1262 (2017)

\bibitem{kinectfusion}
Newcombe, R.A., Izadi, S., Hilliges, O., Molyneaux, D., Kim, D., Davison, A.J., Kohi, P., Shotton, J., Hodges, S., Fitzgibbon, A.: Kinectfusion: Real-time dense surface mapping and tracking. In: 2011 10th IEEE international symposium on mixed and augmented reality. pp. 127--136. Ieee (2011)

\bibitem{nieto2007recursive}
Nieto, J., Bailey, T., Nebot, E.: Recursive scan-matching slam. Robotics and Autonomous systems  \textbf{55}(1),  39--49 (2007)

\bibitem{palieri2020locus}
Palieri, M., Morrell, B., Thakur, A., Ebadi, K., Nash, J., Chatterjee, A., Kanellakis, C., Carlone, L., Guaragnella, C., Agha-Mohammadi, A.a.: Locus: A multi-sensor lidar-centric solution for high-precision odometry and 3d mapping in real-time. IEEE Robotics and Automation Letters  \textbf{6}(2),  421--428 (2020)

\bibitem{prisacariu2017infinitam}
Prisacariu, V.A., K{\"a}hler, O., Golodetz, S., Sapienza, M., Cavallari, T., Torr, P.H., Murray, D.W.: Infinitam v3: A framework for large-scale 3d reconstruction with loop closure. arXiv preprint arXiv:1708.00783  (2017)

\bibitem{qin2018vins}
Qin, T., Li, P., Shen, S.: Vins-mono: A robust and versatile monocular visual-inertial state estimator. IEEE Transactions on Robotics  \textbf{34}(4),  1004--1020 (2018)

\bibitem{pointslam}
Sandstr{\"o}m, E., Li, Y., Van~Gool, L., Oswald, M.R.: Point-slam: Dense neural point cloud-based slam. In: Proceedings of the IEEE/CVF International Conference on Computer Vision. pp. 18433--18444 (2023)

\bibitem{segal2009generalized}
Segal, A., Haehnel, D., Thrun, S.: Generalized-icp. In: Robotics: science and systems. vol.~2, p.~435. Seattle, WA (2009)

\bibitem{serafin2015nicp}
Serafin, J., Grisetti, G.: Nicp: Dense normal based point cloud registration. In: 2015 IEEE/RSJ International Conference on Intelligent Robots and Systems (IROS). pp. 742--749. IEEE (2015)

\bibitem{shan2018lego}
Shan, T., Englot, B.: Lego-loam: Lightweight and ground-optimized lidar odometry and mapping on variable terrain. In: 2018 IEEE/RSJ International Conference on Intelligent Robots and Systems (IROS). pp. 4758--4765. IEEE (2018)

\bibitem{shan2020lio}
Shan, T., Englot, B., Meyers, D., Wang, W., Ratti, C., Rus, D.: Lio-sam: Tightly-coupled lidar inertial odometry via smoothing and mapping. In: 2020 IEEE/RSJ international conference on intelligent robots and systems (IROS). pp. 5135--5142. IEEE (2020)

\bibitem{straub2019replica}
Straub, J., Whelan, T., Ma, L., Chen, Y., Wijmans, E., Green, S., Engel, J.J., Mur-Artal, R., Ren, C., Verma, S., et~al.: The replica dataset: A digital replica of indoor spaces. arXiv preprint arXiv:1906.05797  (2019)

\bibitem{sturm2012benchmark}
Sturm, J., Engelhard, N., Endres, F., Burgard, W., Cremers, D.: A benchmark for the evaluation of rgb-d slam systems. In: 2012 IEEE/RSJ international conference on intelligent robots and systems. pp. 573--580. IEEE (2012)

\bibitem{imap}
Sucar, E., Liu, S., Ortiz, J., Davison, A.J.: imap: Implicit mapping and positioning in real-time. In: Proceedings of the IEEE/CVF International Conference on Computer Vision. pp. 6229--6238 (2021)

\bibitem{taud2018multilayer}
Taud, H., Mas, J.: Multilayer perceptron (mlp). Geomatic approaches for modeling land change scenarios pp. 451--455 (2018)

\bibitem{vizzo2023kiss}
Vizzo, I., Guadagnino, T., Mersch, B., Wiesmann, L., Behley, J., Stachniss, C.: Kiss-icp: In defense of point-to-point icp--simple, accurate, and robust registration if done the right way. IEEE Robotics and Automation Letters  \textbf{8}(2),  1029--1036 (2023)

\bibitem{whelan2013robust}
Whelan, T., Johannsson, H., Kaess, M., Leonard, J.J., McDonald, J.: Robust real-time visual odometry for dense rgb-d mapping. In: 2013 IEEE International Conference on Robotics and Automation. pp. 5724--5731. IEEE (2013)

\bibitem{wu20234d}
Wu, G., Yi, T., Fang, J., Xie, L., Zhang, X., Wei, W., Liu, W., Tian, Q., Wang, X.: 4d gaussian splatting for real-time dynamic scene rendering. arXiv preprint arXiv:2310.08528  (2023)

\bibitem{gsslam}
Yan, C., Qu, D., Wang, D., Xu, D., Wang, Z., Zhao, B., Li, X.: Gs-slam: Dense visual slam with 3d gaussian splatting. arXiv preprint arXiv:2311.11700  (2023)

\bibitem{voxfusion}
Yang, X., Li, H., Zhai, H., Ming, Y., Liu, Y., Zhang, G.: Vox-fusion: Dense tracking and mapping with voxel-based neural implicit representation. In: 2022 IEEE International Symposium on Mixed and Augmented Reality (ISMAR). pp. 499--507. IEEE (2022)

\bibitem{zhang2014loam}
Zhang, J., Singh, S.: Loam: Lidar odometry and mapping in real-time. In: Robotics: Science and systems. vol.~2, pp.~1--9. Berkeley, CA (2014)

\bibitem{nice-slam}
Zhu, Z., Peng, S., Larsson, V., Xu, W., Bao, H., Cui, Z., Oswald, M.R., Pollefeys, M.: Nice-slam: Neural implicit scalable encoding for slam. In: Proceedings of the IEEE/CVF Conference on Computer Vision and Pattern Recognition. pp. 12786--12796 (2022)

\end{thebibliography}
}
\end{document}